\documentclass{article}
\usepackage{spconf,amsmath,graphicx}
\usepackage[normalem]{ulem}

\usepackage{hyperref}

\usepackage{enumitem}
\setlist{nosep, leftmargin=14pt}
\usepackage{booktabs}
\usepackage{mwe} % to get dummy images

\title{SA-UNetv2: Rethinking Spatial Attention U-Net for Retinal Vessel Segmentation}
\name{\parbox{\textwidth}{\centering
Changlu Guo$^{1}$\thanks{Email: \{chagu,anym,abda,mohan\}@dtu.dk; yiyg510@jxnu.edu.cn}, 
Anders Nymark Christensen$^{1}$, 
Anders Bjorholm Dahl$^{1}$, 
Yugen Yi$^{2}$, 
Morten Rieger Hannemose$^{1}$}}

\address{$^{1}$Department of Applied Mathematics and Computer Science,\\
Technical University of Denmark, Kgs.\ Lyngby, Denmark \\
$^{2}$School of Artificial Intelligence,\\
Jiangxi Normal University, Nanchang, China}

\begin{document}
%\ninept
%
\maketitle

\begin{abstract}
Retinal vessel segmentation is essential for early diagnosis of diseases such as diabetic retinopathy, hypertension, and neurodegenerative disorders. Although SA-UNet introduces spatial attention in the bottleneck, it underuses attention in skip connections and does not address the severe foreground–background imbalance. We propose SA-UNetv2, a lightweight model that injects cross-scale spatial attention into all skip connections to strengthen multi-scale feature fusion and adopts a weighted Binary Cross-Entropy (BCE) + Matthews Correlation Coefficient (MCC) loss to improve robustness to class imbalance. On the public DRIVE and STARE datasets, SA-UNetv2 achieves state-of-the-art performance with only \textbf{1.2MB} memory and \textbf{0.26M} parameters (less than 50\% of SA-UNet), and \textbf{1 second} CPU inference on $592 \times 592 \times 3$ images, demonstrating strong efficiency and deployability in resource-constrained, CPU-only settings. \href{https://github.com/clguo/SA-UNetv2}{The code is available at \texttt{github.com/clguo/SA-UNetv2}}.

\end{abstract}

\section{Introduction}
% The morphological structure of retinal vasculature is crucial for the early diagnosis of systemic diseases such as diabetes and hypertension. Pathologies like Diabetic Retinopathy (DR) and Hypertensive Retinopathy (HR) often involve dilation, tortuosity, or occlusion of fine vessels. Accurate vessel segmentation in fundus images enables the extraction of geometric features—such as diameter, curvature, and connectivity—which supports disease monitoring and biometric applications. However, the complex vascular morphology, low contrast, uneven illumination, and significant class imbalance (with vessel pixels typically $<10\%$ of the image) pose serious challenges to conventional segmentation methods, which often fail to capture thin, low-contrast vessels effectively.

Retinal vasculature is vital for early detection of systemic diseases such as diabetes and hypertension, where fine vessels often exhibit abnormalities like dilation, tortuosity, or occlusion. Accurate segmentation enables extraction of geometric features essential for disease monitoring and biometrics, but remains difficult due to complex branching, low contrast from uneven illumination and noise, and severe class imbalance (vessel pixels typically ${<}$10\%)~\cite{kande2025multi}.

\begin{figure}[t]
  \centering
   \includegraphics[width=1\linewidth]{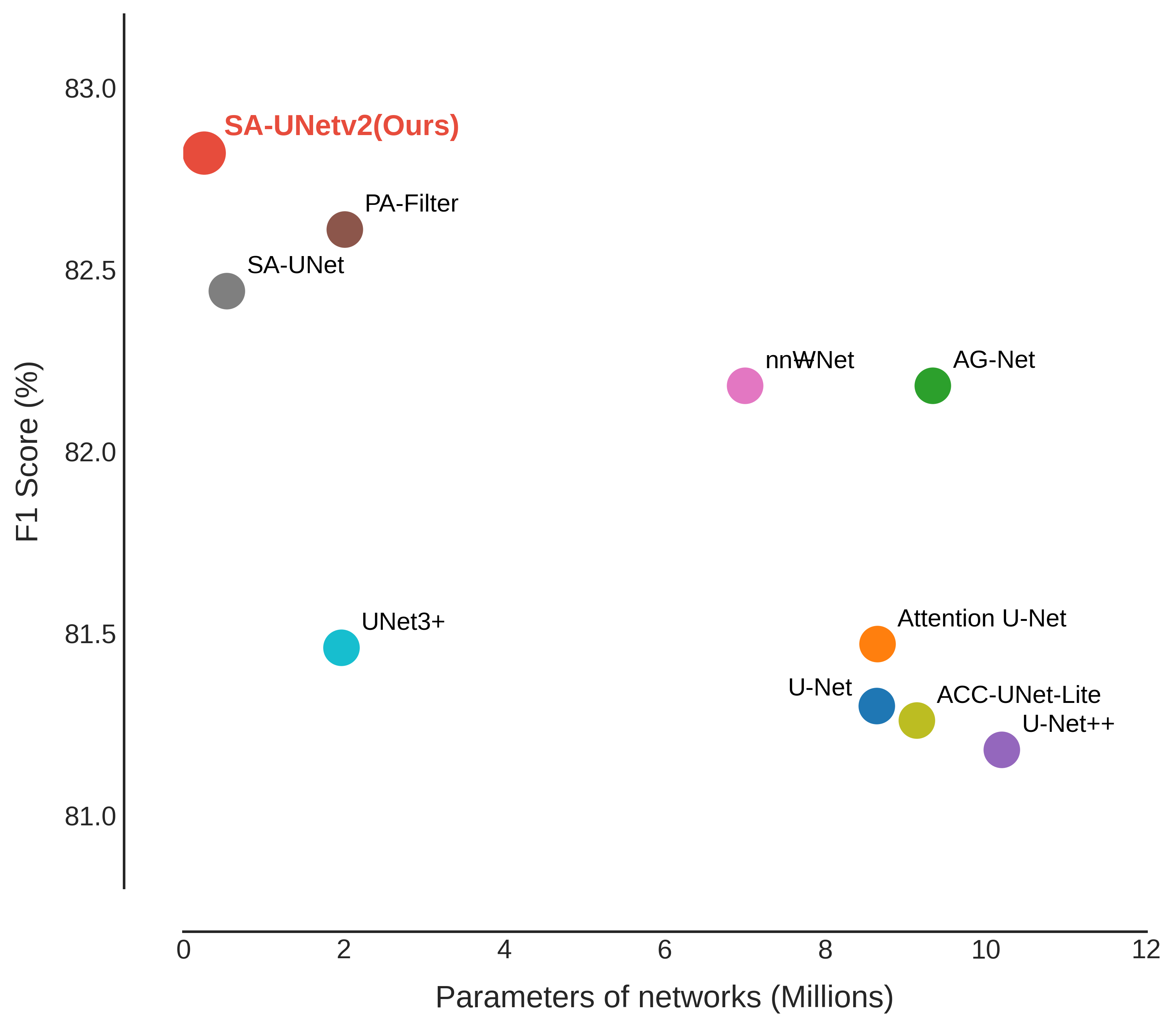}

   \caption{Comparison of retinal vessel segmentation networks on the DRIVE dataset. Our SA-UNetv2 achieves the highest F1 score with the lowest model complexity.  }
   \label{fig:pars}
\end{figure}

With the rapid advancement of deep learning, such as U-Net~\cite{ronneberger2015u} and its variants including Attention U-Net~\cite{oktay2018attention}, U-Net++~\cite{zhou2019unet++}, U-Net 3+~\cite{huang2020unet}, ACC-UNet~\cite{ibtehaz2023acc}, and U-Netv2~\cite{peng2025u} have become the de facto standard for medical image segmentation. In retinal vessel segmentation, numerous tailored models have been introduced, including AG-Net~\cite{zhang2019attention}, IterNet~\cite{li2020iternet}, MamUNet~\cite{han2025mamunet}, and others~\cite{qin2024review}. However, their large parameter sizes (9.34M, 8.25M, and 16.86M) make deployment on CPU-based or resource-limited systems challenging. Although RetinalLiteNet~\cite{mehmood2024retinalitenet} reduces the model size to 0.066M parameters, this comes at the expense of segmentation performance. SA-UNet~\cite{guo2021sa}, our earlier work, a 0.54M model that integrates spatial attention in the bottleneck and employs DropBlock~\cite{ghiasi2018dropblock} with Batch Normalization to achieve a better balance between performance and efficiency. However, SA-UNet still presents two notable limitations: (1) spatial attention is confined to the bottleneck, restricting multi-scale feature fusion that is crucial for capturing fine vessel structures; and (2) relying solely on Binary Cross-Entropy (BCE) loss inadequately addresses the severe vessel–background imbalance, leading to a bias toward background pixels and reduced sensitivity to thin, low-contrast vessels. To overcome these challenges, we propose SA-UNetv2, an enhanced and more lightweight architecture specifically tailored for retinal vessel segmentation. It introduces two key innovations: (1) a novel Cross-scale Spatial Attention (CSA) module that integrates attention across encoder and decoder pathways, effectively bridging the semantic gap and improving the detection of fine vessels; and (2) a compound loss function that combines BCE with Matthews Correlation Coefficient (MCC) loss, enhancing robustness to severe class imbalance by optimizing segmentation from both local and global perspectives. Experiments conducted on two public benchmarks, DRIVE and STARE, demonstrate that SA-UNetv2 achieves state-of-the-art segmentation performance while being one of the most lightweight architectures among existing methods, as shown in Fig. \ref{fig:pars}. Notably, it delivers sub-second inference even on CPU, making it a practical and efficient solution for real-world retinal image analysis.

\begin{figure}[t]
  \centering
   \includegraphics[width=1\linewidth]{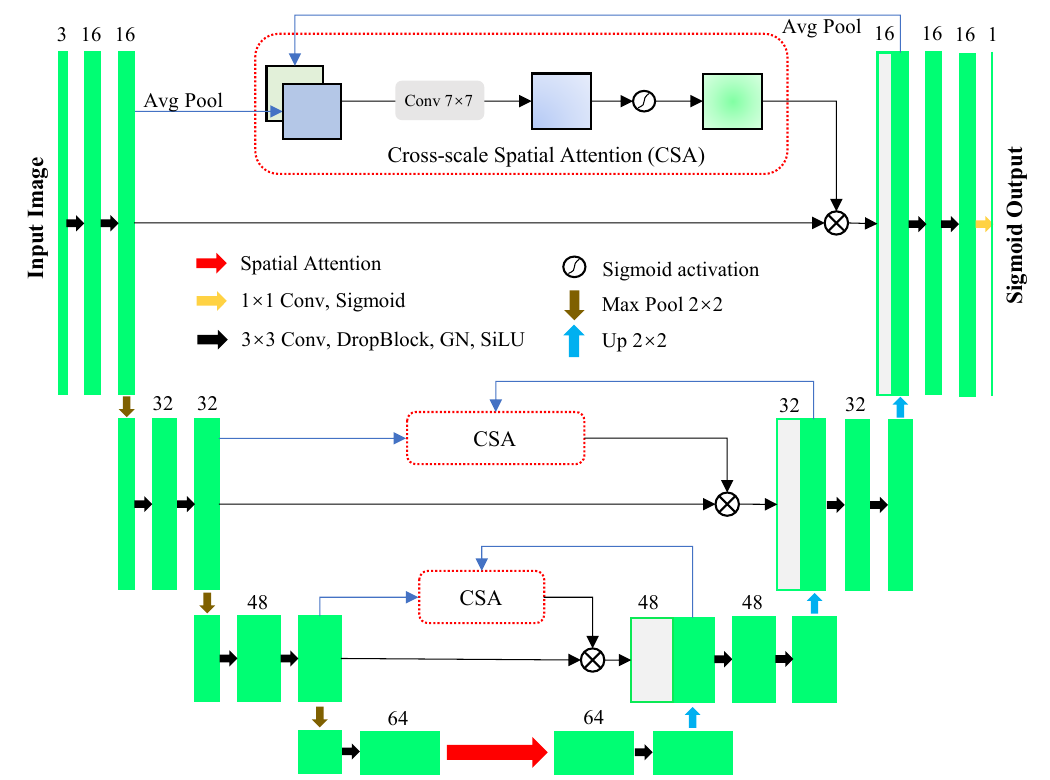}

   \caption{The  Architecture of SA-UNetv2.
   }
   \label{fig:saunetv2}
\end{figure}

\section{SA-UNetV2}

\subsection{Architecture}
SA-UNetv2 builds upon the simplicity and efficiency of the original SA-UNet while addressing its limitations in convolutional block design, parameter efficiency, and skip-connection feature fusion through three structural enhancements, as illustrated in Fig.~\ref{fig:saunetv2}. First, the core convolutional unit is redesigned from the conventional \textit{Conv 3×3 → DropBlock → Batch Normalization → ReLU} sequence to an optimized \textit{Conv 3×3 → DropBlock → Group Normalization → SiLU} structure, where Group Normalization (GN)~\cite{wu2018group} removes the dependency on batch size, making the network more stable for small-batch medical image training, and the smooth nonlinearity of SiLU~\cite{elfwing2018sigmoid} improves gradient flow, aiding in the capture of subtle structures in low-contrast lesions. Second, the feature channel configuration is compressed from [16, 32, 64, 128] to a more parameter-efficient [16, 32, 48, 64], reducing the parameter count from 0.54M to 0.26M (a reduction of approximately 51.9\%) while maintaining multi-scale discriminative capacity, thus improving deployment efficiency. Finally, and most importantly, we propose the Cross-scale Spatial Attention (CSA) module and, for the first time in this architecture lineage, integrate it into all skip connections to explicitly address the semantic gap between encoder and decoder features in U-Net. In contrast to the original SA module, which generates attention maps solely from encoder features, CSA jointly exploits complementary information from both the encoder feature $F^{e}$ and the decoder feature $F^{d}$. By incorporating decoder context, CSA enables bidirectional feature interaction that effectively bridges the encoder–decoder semantic gap. Specifically, $F^{e}$ and $F^{d}$ are separately processed by channel-wise average pooling, concatenated, and passed through a $7\times 7$ convolution followed by a sigmoid activation to produce a cross-scale spatial attention map:
\[
F^{out} = F^{e} \cdot \sigma\!\left(f^{7\times 7}\big[\text{AvgPool}(F^{e});\,\text{AvgPool}(F^{d})\big]\right),
\]
where $f^{7\times7}(\cdot)$ denotes the $7\times7$ convolution, $\sigma(\cdot)$ the sigmoid function, and $\text{AvgPool}(\cdot)$ the average pooling. With only 98 parameters—identical to the original SA module—CSA effectively enhances cross-scale feature fusion at 32-, 48-, and 64-channel skip connections, improving fine-grained structural delineation and global semantic consistency.

\subsection{Loss Function Design}
Retinal vessel segmentation faces extreme class imbalance, with vessel pixels typically occupying less than 10\% of the image area. In such scenarios, Binary Cross-Entropy (BCE)
\begin{equation}
\mathcal{L}_{\mathrm{BCE}} = -\frac{1}{N}\sum_{i=1}^N \big[ y_i\log p_i + (1-y_i)\log(1-p_i) \big]
\end{equation}
is dominated by background pixels, leading to low sensitivity for thin vessels, where $y_i \in \{0,1\}$ denotes the ground-truth label of pixel $i$, $p_i \in [0,1]$ is the predicted probability, and $N$ is the total number of pixels. To address this limitation, we introduce a differentiable Matthews Correlation Coefficient (MCC) loss:
\begin{equation}
\resizebox{\columnwidth}{!}{$
\mathcal{L}_{\mathrm{MCC}} = 1 -
\frac{\mathrm{TP}\cdot\mathrm{TN} - \mathrm{FP}\cdot\mathrm{FN}}
{\sqrt{(\mathrm{TP}+\mathrm{FP})(\mathrm{TP}+\mathrm{FN})
(\mathrm{TN}+\mathrm{FP})(\mathrm{TN}+\mathrm{FN})} + \epsilon}
$}
\end{equation}
where $\mathrm{TP}=\sum p_i y_i$, $\mathrm{TN}=\sum (1-y_i)(1-p_i)$, $\mathrm{FP}=\sum (1-y_i)p_i$, and $\mathrm{FN}=\sum y_i(1-p_i)$. By preserving continuous probabilities $p_i$ without thresholding, using only differentiable operations (addition, multiplication, square root), and adding $\epsilon=10^{-7}$ for numerical stability, $\mathcal{L}_{\mathrm{MCC}}$ remains fully end-to-end differentiable. The total loss is defined as
$
\mathcal{L}_{\mathrm{total}} = \lambda_1\,\mathcal{L}_{\mathrm{BCE}} + \lambda_2\,\mathcal{L}_{\mathrm{MCC}},
$
where $\lambda_1$ and $\lambda_2$ are non-negative weights balancing the BCE and MCC terms. This formulation enforces pixel-level accuracy and global consistency, enhancing sensitivity while preserving high specificity.

\begin{table*}[htbp]
  \centering
  \fontsize{8pt}{8pt}\selectfont
  \caption{Comparison of Methods on DRIVE Datasets. ($^\dagger$ indicates results reported from MamUNet \cite{han2025mamunet}) }
  \resizebox{\textwidth}{!}{
    \begin{tabular}{l l ccccccccccc}
      \toprule
      Model & Publication & F1 & Jacc & Sen & Spe & ACC & MCC & AUC & GFLOPs & Pars (M) & Mem(MB) & Time(s) \\
      \midrule
      \multicolumn{13}{c}{Without FOV Mask} \\
      \midrule
      U-Net \cite{ronneberger2015u}        & MICCAI2015 & 81.30 & 68.52 & 80.65 & 98.34 & 96.77 & 79.66 & 98.24 & 137.10 & 8.64 & 34.72 & 3.16 \\
      Attention U-Net \cite{oktay2018attention} & MIDL2018   & 81.47 & 68.76 & 81.24 & 98.30 & 96.78 & 79.84 & 98.22 & 425.05 & 8.65 & 34.80 & 7.74 \\
      AG-Net \cite{zhang2019attention}     & MICCAI2019 & --    & 69.65 & 81.00 & \textbf{98.48} & 96.92 & 79.84 & 98.56 & --    & 9.34 & 37.36 & --   \\
      U-Net++  \cite{zhou2019unet++}       & TMI2020    & 81.18 & 68.34 & 82.40 & 98.07 & 96.67 & 79.50 & 98.44 & 338.25 & 10.20 & 41.08 & 6.95 \\
      PA-Filter  \cite{li2022retinal}      & ISBI2020   & 82.61 & --    & --    & --    & \textbf{96.99} & --    & 98.43 & --    & 2.01  & --    & --   \\
      UNet3+ \cite{huang2020unet}          & ICASSP2020 & 81.46 & 68.74 & 82.02 & 98.18 & 96.75 & 79.79 & 98.47 & 198.20 & 1.97 & 8.36  & 5.29 \\
      ACC-UNet-Lite \cite{ibtehaz2023acc}  & MICCAI2023 & 81.26 & 68.47 & 82.05 & 98.14 & 96.71 & 79.57 & 98.34 & 215.59 & 9.14 & 37.05 & 4.41 \\
      nn{\sout{W}}Net \cite{zhou2025nnwnet}& CVPR2025   & 82.18 & 69.86 & --    & --    & --    & --    & --    & --     & 7.00 & --    & --   \\
      SA-UNet \cite{guo2021sa}             & ICPR2020   & 82.44 & 70.15 & 83.64 & 98.19 & 96.90 & 80.85 & 98.62 & 26.54  & 0.54 & 2.29  & 1.12 \\
      SA-UNetv2                            & --         & \textbf{82.82} & \textbf{70.69} & \textbf{83.64} & 98.28 & 96.98 & \textbf{81.27} & \textbf{98.71} & \textbf{21.19} & \textbf{0.26} & \textbf{1.20} & \textbf{0.95} \\
      \midrule
      \multicolumn{13}{c}{With FOV Mask} \\
      \midrule
      IterNet \cite{li2020iternet}         & WACV2020   & 82.18 & --    & 77.91 & \textbf{98.31} & \textbf{95.74} & --    & \textbf{98.13} & --     & 8.25  & --    & --   \\
      RetinalLiteNet \cite{mehmood2024retinalitenet} & CVPRW2024 & 80.60 & 67.50 & 78.40 & 98.00 & --    & --    & 97.00 & --     & \textbf{0.066} & \textbf{0.25} & --   \\
      UNetv2$^\dagger$ \cite{peng2025u}              & ISBI2025   & 79.64 & --    & --    & --    & 94.72 & --    & 87.62 & --     & 25.15 & --    & --   \\
      MamUNet \cite{han2025mamunet}        & ISBI2025   & 81.78 & --    & --    & --    & 95.36 & --    & 90.25 & --     & 16.86 & --    & --   \\
      SA-UNet \cite{guo2021sa}             & ICPR2020   & 82.46 & 70.18 & 83.67 & 97.25 & 95.49 & 80.00 & 97.94 & 26.54  & 0.54 & 2.29  & 1.12 \\
      SA-UNetv2                            & --         & \textbf{82.84} & \textbf{70.73} & \textbf{83.67} & 97.39 & 95.61 & \textbf{80.44} & 98.08 & \textbf{21.19} & 0.26 & 1.20  & \textbf{0.95} \\
      \bottomrule
    \end{tabular}
  }
  \label{tab:drive}
\end{table*}

\begin{table}[htbp]
  \centering
  \fontsize{8pt}{8pt}\selectfont
  \caption{Comparison of Methods on the \textsc{STARE} Dataset.}
  \resizebox{\columnwidth}{!}{%
    \begin{tabular}{l ccccccc}
      \toprule
      Model & F1 & Jacc & Sen & Spe & ACC & MCC & AUC \\
      \midrule
      IterNet \cite{li2020iternet} & 81.46&--&77.15&\textbf{99.19}&97.82&--&99.15 \\
      
      U\textendash Net  \cite{ronneberger2015u}          & 79.74 & 66.43 & 83.88 & 98.45 & 97.45 & 78.77 & 98.90 \\
      Attention U\textendash Net \cite{oktay2018attention} & 80.61 & 67.72 & 84.63 & 98.49 & 97.55 & 79.60 & 98.46 \\
      U\textendash Net++  \cite{zhou2019unet++}       & 79.56 & 66.15 & 79.45 &98.82 & 97.53 & 78.49 & 98.82 \\

      PA-Filter   \cite{li2022retinal} &81.70&--&--&--&\textbf{97.88}&--&98.43 \\
      UNet3+    \cite{huang2020unet}                 & 81.16 & 68.40 & 84.50 & 98.60 & 97.60 & 80.34 & 99.06 \\
      ACC\textendash UNet\textendash Lite \cite{ibtehaz2023acc} & 78.99& 65.47 & 86.12 & 98.12 & 97.24 & 78.30 & 98.85 \\

      SA\textendash UNet \cite{guo2021sa} & 80.84 & 68.01 & \textbf{89.99} & 98.03 & 97.45 & 80.19 & \textbf{99.18} \\

      SA\textendash UNetv2       & \textbf{82.81} & \textbf{70.82} & 85.35 & 98.71 & 97.83 & \textbf{81.79} & 99.13 \\
      \bottomrule
    \end{tabular}%
  }
  \label{tab:stare_results}
\end{table}

\section{Experiments}
\label{sec:Experiments}

\subsection{Datasets and Setup}
This study employs two publicly available datasets widely used for retinal vessel segmentation tasks---DRIVE and STARE. The DRIVE dataset, obtained from the Dutch Diabetic Retinopathy Screening Program, contains 40 color fundus images, with 20 images for training and 20 for testing. The STARE (Structured Analysis of the Retina) dataset, developed by the University of Florida, comprises 20 fundus images with a resolution of 700$\times$605 pixels. Since the STARE dataset does not provide an official training/testing split, we follow prior literature \cite{li2020iternet} \cite{li2022retinal} by using the first 16 images for training and the remaining 4 images for testing. To ensure consistent input sizes during training, we follow the protocol and data augmentation of SA-UNet~\cite{guo2021sa}: DRIVE images (584$\times$565) are zero-padded to 592$\times$592, and STARE images (700$\times$605) to 704$\times$704, with 10\% of the augmented training data randomly selected for validation. During testing, the model outputs are cropped back to their original sizes (DRIVE: 584$\times$565, STARE: 700$\times$605) to guarantee accurate and comparable evaluation. As the STARE dataset does not provide an official field-of-view (FOV) mask, all results on STARE are reported without FOV masking.

To ensure a fair comparison, we trained several classical U-Net variants from scratch, including U-Net \cite{ronneberger2015u} , Attention U-Net \cite{oktay2018attention}, U-Net++ \cite{zhou2019unet++} , UNet3+ \cite{huang2020unet} , ACC-UNet-Lite \cite{ibtehaz2023acc} , and SA-UNet \cite{guo2021sa}, all adopting an identical convolutional block structure \textit{Conv 3×3 → DropBlock → Group Normalization → SiLU}. This setup guarantees architectural consistency across models while leveraging the proven regularization capability of DropBlock in SA-UNet to effectively mitigate overfitting in U-Net-based architectures. All models were trained under identical settings using the Adam optimizer with an initial learning rate of \(1\times10^{-3}\), a composite loss function of \(0.5 \times \,\mathrm{BCE} + 0.5 \times \,\mathrm{MCC}\), and DropBlock regularization (dropout rate of 0.15, block size of 7). Training was performed for a maximum of 150 epochs with an early stopping strategy to prevent overfitting. The batch size was set to 8 for the DRIVE dataset and 2 for the STARE dataset.

The computational complexity of each model was evaluated using the \texttt{keras\_flops} tool, calculating floating-point operations (GFLOPs) based on an input size of \(592\times592\times3\). The number of trainable parameters (Pars) and the model memory footprint (Mem) were also reported to indicate model scale. Evaluation metrics included F1 score, Jaccard index (Jacc), sensitivity (Sen), specificity (Spe), accuracy (ACC), Matthews correlation coefficient (MCC), and the area under the ROC curve (AUC). Inference efficiency was measured on a Kaggle Intel(R) Xeon(R) CPU @ 2.20\,GHz (2 cores, 4 threads) with batch size 1, as the average time per image over 20 inputs of \(592\times592\times3\).

\subsection{Results}
\begin{figure}[t]
  \centering
   \includegraphics[width=1\linewidth]{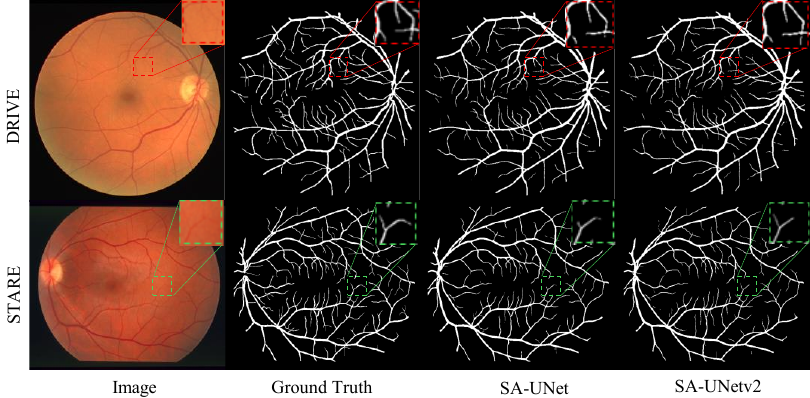}

   \caption{Segmentation comparison: SA-UNet vs. SA-UNetv2.
   }
   \label{fig:results}
\end{figure}

\textbf{Comparison with State-of-the-Art Methods
} In addition to models we re-trained from scratch (e.g., U-Net and its variants~\cite{ronneberger2015u,oktay2018attention,zhou2019unet++,huang2020unet,ibtehaz2023acc,guo2021sa}), we also include results reported in recent works~\cite{peng2025u,zhang2019attention,li2020iternet,han2025mamunet,mehmood2024retinalitenet,li2022retinal,chen2024hidiff,zhou2025nnwnet} for reference. As shown in Tables~\ref{tab:drive} and~\ref{tab:stare_results}, SA-UNetv2 achieves leading performance in both accuracy and efficiency. Under the primary evaluation protocol on DRIVE without FOV, it attains the highest scores across key metrics: F1 (82.82), Jaccard (70.69), MCC (81.27), and AUC (98.71). Compared with the previous best-performing methods---PA-Filter (F1 82.61) and SA-UNet (Jaccard 70.15)---SA-UNetv2 improves F1 by +0.21 and Jaccard by +0.54, respectively. Relative to the SA-UNet baseline (26.54~GFLOPs, 0.54M parameters, 2.29~MB model, 1.12s CPU inference), SA-UNetv2 reduces GFLOPs to 21.19, halves parameters to 0.26M, compresses model size to 1.20~MB, and shortens inference to 0.95s, while further improving accuracy. We also report DRIVE results with FOV for comparison with the most lightweight method, RetinalLiteNet~\cite{mehmood2024retinalitenet}, which contains only 0.066M parameters and a 0.25~MB file but achieves limited accuracy (F1 80.60, Jaccard 67.50). Within a similar low-complexity range, SA-UNetv2 (0.26M parameters) delivers markedly better segmentation (F1 82.84, Jaccard 70.73, MCC 80.44, AUC 98.08), improving F1 by +2.24 and Jaccard by +3.23 over RetinalLiteNet. These results demonstrate that SA-UNetv2 achieves a well-balanced state-of-the-art performance, maintaining sub-million parameters and low GFLOPs while significantly enhancing segmentation accuracy. On the STARE dataset, SA-UNetv2 again leads across the board, achieving F1 of 82.81, Jaccard of 70.82, and MCC of 81.79. Compared to the best previously reported results---PA-Filter (F1 81.70) and UNet3+ (Jaccard 68.40)---SA-UNetv2 improves F1 by +1.11 and Jaccard by +2.42, demonstrating robust generalization across datasets and evaluation protocols. Moreover, as illustrated in Fig.~\ref{fig:results}, SA-UNetv2 also exhibits superior capability in delineating fine-grained vascular structures compared with SA-UNet, particularly in the segmentation of thin and low-contrast vessels. Overall, SA-UNetv2 delivers clear gains in accuracy over prior state-of-the-art methods and significant improvements over prior lightweight models, while keeping parameters and GFLOPs low.
\begin{table}[htbp]
  \centering
  \fontsize{8pt}{8pt}\selectfont   % 固定字体大小为 8pt，行距 9pt
  \setlength{\tabcolsep}{2pt}      % 设置列间距（默认约 6pt，可改成 3~5pt 之间）
  \caption{Ablation Study on DRIVE (Only BCE Loss).}
  \resizebox{\columnwidth}{!}{%
    \begin{tabular}{lccccccc}
      \toprule
      Method & Channels & F1 & Jacc & MCC & ACC & Pars (M) & GFLOPs \\
      \midrule
      SA-UNet (BN,ReLU)               & (16,32,64,128)      & 82.17 & 69.77 & 80.67 & 96.96 & 0.54  & 26.30 \\
      SA-UNet (GN,SiLU)               & (16,32,64,128)      & 82.49 & 70.22 & 80.92 & 96.92 & 0.54  & 26.54 \\
      SA-UNet (GN,SiLU)               & (16,32,48,64)       & 82.70 & 70.52 & 81.15 & 96.97 & 0.26  & 21.19 \\
      SA-UNet (GN,SiLU) + SA          & (16,32,48,64)       & 82.65 & 70.45 & 81.11 & 96.97 & 0.26  & 21.19 \\
      SA-UNetv2            & (16,32,48,64)       & \textbf{82.75} & \textbf{70.60} & \textbf{81.21} & \textbf{96.99} & 0.26  & 21.19 \\
      \bottomrule
    \end{tabular}%
  }
  \label{tab:tab2}
\end{table}

\noindent \textbf{Component Analysis} Since SA-UNet, compared to a U-Net with the same channel configuration (16, 32, 64, 128), has already demonstrated the effectiveness of DropBlock and the SA module in the bottleneck, this study directly adopts SA-UNet as the baseline for ablation (Table~\ref{tab:tab2}), using BCE as the unified loss. Replacing BN and ReLU with GN and SiLU increases F1 from 82.17 to 82.49 and Jaccard from 69.77 to 70.22, validating the impact of improved normalization and activation. Reducing the channel configuration from (16, 32, 64, 128) to (16, 32, 48, 64) halves the parameters while slightly improving performance, indicating that a lightweight design retains sufficient feature capacity. Introducing the SA module into skip connections does not yield further gains, suggesting this location is unsuitable—likely due to the semantic gap between encoder and decoder—thus confirming the original design. Ultimately, SA-UNetv2 integrates GN, SiLU, the compact configuration, and the CSA module, achieving the best performance (F1 82.75, Jaccard 70.60, MCC 81.21) with only 0.26M parameters and 21.19 GFLOPs, striking a balance between accuracy and efficiency.

\begin{table}[htbp]
  \centering
  \fontsize{8pt}{9pt}\selectfont    % 固定字体大小
  \setlength{\tabcolsep}{4pt}       % 控制列间距
  \caption{SA-UNetv2 with Different Loss Functions.}
  \resizebox{\columnwidth}{!}{%
    \begin{tabular}{lccccccc}
      \toprule
      Loss Function (Weight) & F1 & Jacc & Sen & Spe & ACC & MCC & AUC \\
      \midrule
      BCE (1.0)            & 82.75 & 70.60 & 82.81 & \textbf{98.38} & \textbf{96.99} & 81.21 & 98.70 \\
      MCC (1.0)            & 82.73 & 70.56 & 84.35 & 98.16 & 96.93 & 81.17 & 91.71 \\
      BCE+MCC (0.7,0.3)    & 82.62 & 70.41 & \textbf{84.85} & 98.07 & 96.89 & 81.06 & 98.71 \\
      BCE+MCC (0.5,0.5)    & \textbf{82.82} & \textbf{70.69} & 83.64 & 98.28 & 96.98 & \textbf{81.27} & \textbf{98.71} \\
      BCE+MCC (0.3,0.7)    & 82.76 & 70.61 & 83.81 & 98.25 & 96.96 & 81.21 & 98.68 \\
      \bottomrule
    \end{tabular}%
  }
  \label{tab:loss_ablation}
\end{table}
\noindent \textbf{Loss Function Analysis} We further evaluate the effect of different loss combinations on DRIVE, as shown in Table~\ref{tab:loss_ablation}.. BCE alone achieves strong overall results (F1 82.75, ACC 96.99) and highest specificity (98.38). In contrast, MCC alone slightly lowers ACC and AUC but improves sensitivity (84.35), demonstrating its advantage in capturing minority class features. Combining BCE and MCC with equal weighting (0.5:0.5) yields the best overall performance, with the highest F1 (82.82), Jaccard (70.69), and MCC (81.27), while maintaining a good balance between sensitivity and specificity. Therefore, SA-UNetv2 adopts the equally weighted BCE+MCC loss to ensure stable optimization and consistently strong performance across multiple metrics.

\section{Conclusion}

In summary, SA-UNetv2, with only 0.26M parameters, integrates cross-scale spatial attention and enhanced convolutional blocks to achieve state-of-the-art performance on both DRIVE and STARE, delivering superior accuracy–efficiency trade-offs with sub-second CPU inference on DRIVE, and future work will focus on extending its adaptability to other retinal imaging modalities and real-time clinical screening.

\bibliographystyle{IEEEbib}
\bibliography{refs}

\end{document}